# Recent Advances in Content Based Video Copy Detection

S. R. Shinde
PG Student, Dept. of Computer Engineering
Sinhgad College of Engineering
Pune, India
r3t_sanket@rediffmail.com

G. G. Chiddarwar
Assistant Professor, Dept. of Computer Engineering
Sinhgad College of Engineering
Pune, India
ggchiddarwar.scoe@sinhgad.edu

*Abstract*—With the immense number of videos being uploaded to the video sharing sites, issue of copyright infringement arises with uploading of illicit copies or transformed versions of original video. Thus safeguarding copyright of digital media has become matter of concern. To address this concern, it is obliged to have a video copy detection system which is sufficiently robust to detect these transformed videos with ability to pinpoint location of copied segments. This paper outlines recent advancement in content based video copy detection, mainly focusing on different visual features employed by video copy detection systems. Finally we evaluate performance of existing video copy detection systems.

*Keywords—Copyright protection, content based video copy detection, feature extraction, feature descriptor, MUSCLE-VCD, TRECVID*

## I. INTRODUCTION

The expeditious growth of the World Wide Web has allowed netizens in acquiring and sharing digital media in relatively simpler way due to improvements in data transfer and processing capabilities. Due to wide use of digital devices like smart phones, cameras, more and more images and videos are produced by netizens and are uploaded on the internet for business promotions or community sharing.

The very easiness of video copy creation techniques instigated problem of video copyright violations, so it is needed to have mechanism to protect copyright of digital videos. In September 2014, according to YouTube statistics [1], following facts about viewership came in light,

1. A video sharing site, YouTube has 1 billion unique visitors every month.
2. Visitors watch over 6000 million hours of motion picture every month on YouTube.
3. Video upload rate on YouTube is 100 hours of video per minute.

As here we see, there is a huge human traffic for video sharing sites and large amount of videos are being uploaded on these sites like YouTube, Dailymotion, Google Video, etc. So this poses problems to media broadcasting groups as identifying illicit versions of an original video has become a challenging task.

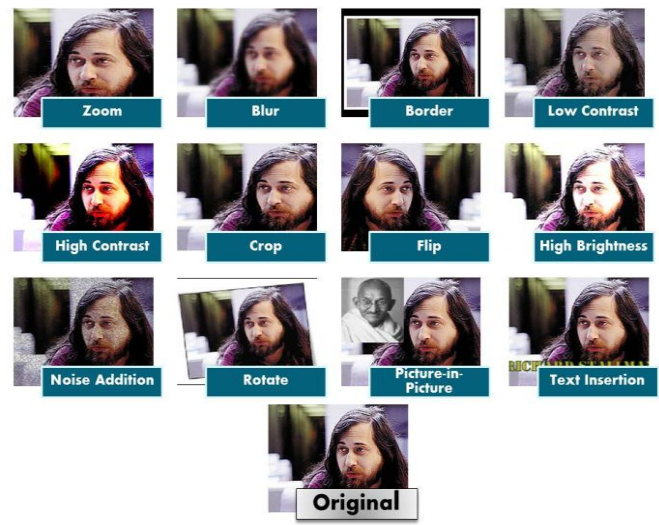

Fig. 1. Common video transformations

Thus video copy detection has become crucial solution to reduce huge piracy and copyright issues.

Existing video copy detection techniques are mainly classified into watermarking based and content based copy detection. Each of these techniques has its own merits and drawbacks. Watermark embeds useful metadata and maintains low computational cost for copy detection operation, but watermark based copy detection does not perform well against transformations like rotate, blur, crop, camcording, resize, which are performed during video copy creation as shown in Fig. 1. If original version of video is distributed on video sharing sites before watermark embedding, then watermark based detection system does not have any reactive measure. Also due to video compression, possibility of vanishing watermark arises.

There are many methods for embedding watermark into an original image. These watermark based schemes are based on fourier, cosine, wavelet transforms. But these transform based methods usually perform embedding of watermark into predefined set of coefficients of their corresponding domain. Thus whenever an attacker scrutinizes image and finds pattern of embedding watermark into predefined set of coefficients, he can easily remove embedded watermark. Another issue is how to decide which set of coefficients to be selected for embedding watermark [2]. If suppose in case of DCT, if we embed watermark into set of coefficients which belongs to high frequency range, then it is quite possible that if low pass filtering attack is implemented then watermark embedded in high frequency coefficients will just vanish. Even if we select low frequency coefficients, in case of DCT, to embed watermark, then it will significantly degrade quality of an image as this comes from fact that a DCT operation on image gives very good energy compaction in the lower frequency region and human vision is able to detect alterations to these frequencies [2].

Recently formulated Content Based Copy Detection (CBCD) algorithms as contrast to watermark-based methods do not rely on any watermark embedding and are invariant to most of the transformations. These CBCD algorithms extract invariant features from the media content itself, so CBCD mechanism can be applied to probe copyright violations of digital media on the internet as an effective alternative to watermarking technique. CBCD algorithms first extract distinct and invariant features from the original and query videos. If same features are found in both original and query videos, then query video may be a copied version of original video. Underlying assumption of CBCD algorithms is that a sufficient amount of information is available in video content itself to generate its unique description; it means content itself preserves its own identity. Although video copy detection issue is perceived as one facet of video retrieval, but basic difference between these two is, video copy detection system finds exact versions of a query video including original and transformed one as shown in Fig. 1, whereas a video retrieval system searches for similar videos.

The crucial issue of copyright infringements has led to much advancement in video copy detection methodologies. Most of the surveys cover only a subset of topics in video copy detection. For example Hampapur et al. [3] evaluated distance/similarity measures used for CBCD implementations; Roopalakshmi et al. [4] illustrated video feature/signature description techniques for CBCD algorithms and briefed research challenges. Bhattacharya et al. [5] gave good review on variety of video watermarking algorithms. J.M. Barrios [6] presented analysis of similarity measures used for matching video sequences. Law-To et al. [7] compared local features with global ones and concluded that copy detection with local features needs more computational time but are highly robust than inexpensive global features. Shiguo Lian et al. [8] investigated video copy detection algorithms through appropriate performance metrics. Hampapur et al. [9] gave a review on different video sequence matching mechanisms used in CBCD systems.

## II. MOTIVATION

As multiple videos are being uploaded on internet either for business promotions or community sharing, many problems gets arise including storage management and copyright violations.

I) First issue is about data redundancy. It is quite expensive to maintain multiple copies of video in a repository as this requires huge storage requirements and causing video retrieval operation more time consuming. If it becomes possible to identify duplicate copies of a video in video repository, then an effective storage management will be achieved.

II) Second issue is related to huge piracy and copyright infringements. Due to easiness in creation of transformed video copy and uploading it on internet, this may cause huge loss for commercial businesses like multimedia groups or broadcasting agencies. As it is not possible for a human operator to go manually through video database to check if any copied version of original video content is present. So these two consequential issues give rise to a need of implementing an automated form of video copy detection system.

Fig. 2 shows general architecture of content based video copy detection system. This system is comprised of mainly two stages; these are elaborated as follows,

1) *Offline stage*: Firstly video preprocessing is done to normalize quality of the video and to eliminate transformation effects as much as possible. Keyframes are extracted from segments of original videos and from every keyframe invariant features are excerpted. These invariant features should be able to detect transformed versions of original video. After feature extraction, features are enlisted into an index data structure to perform faster feature retrieval and matching operations.

2) *Online stage*: In this stage query videos are evaluated. Features extraction is performed on preprocessed keyframes of a query video and extracted features are compared to features stored in an index structure. Then similarity results are examined. Finally system gives copy detection result.

List of video transformations applied to queries by major copy detection datasets is given as below,

1) *MUSCLE-VCD*: This dataset comprised of ground truth data and set of tasks to assess performance of system in copy localization, tasks are: copy detection (ST1) and localizing copy segments from video sequence (ST2). ST1 task includes queries ranging from S1 to S15, some of these are,

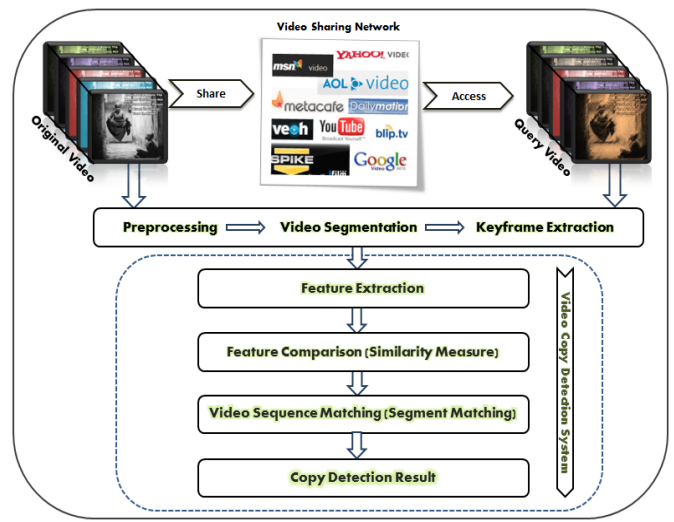

Fig. 2. General architecture of content based video copy detection system

**S1**. *Change of color, blur;* **S3**. *Re-encoding, crop and change of color;* **S5**. *Strong re-encoding;* **S6**. *Camcording, subtitles;* **S9**. *Analogic noise, change in YUV;* **S10**. *Camcording with an angle;* **S11**. *Camcording;* **S13**. *Flip(horizontal mirror);* **S14**. *Zoom, subtitles;* **S15**. *Small resize.*

*2) TRECVID*: This dataset has been changed from time to time based on changes in video transformations. Video queries are generated by applying different photometric and geometric transformations ranging from T1 to T10.

**T1**. *Camcording;* **T2**. *Picture in Picture;* **T3**. *Pattern Insertions;* **T4**. *Strong re-encoding;* **T5**. *Change in gamma;* **T6**. *Any three quality degradations (change in gamma, change of ratio, noise, contrast, blurring, color, frame dropping, change of compression);* **T7**. *Any five quality degradations;* **T8**. *Any three post production transformations (caption, shift, slow motion, flip, crop, picture in picture, contrast);* **T9**. *Any five post production transformations;* **T10**. *Combination of five random transformations.*

This paper is organized as: Section III reviews variety of visual features employed by different video copy detection systems. Table I lists down visual features employed by existing video copy detection systems along with their pros and cons. Section IV evaluates performance by comparative analysis of different visual features. Table II shows detection results of representative video copy detection systems with TRECVID*(2008/2009/2011)* dataset. Finally Section V summarizes this paper.

### III. FEATURE CATEGORIZATION

For attaining both efficiency and effectiveness in video copy detection, the feature signature should adhere to two crucial properties, uniqueness and robustness. Uniqueness stipulates discriminating potential of the feature. While robustness implies potential of noise resistance means features should remain unchanged even in case of different photometric or geometric transformations. Once set of keyframes has been decided, distinct features are extracted from keyframes and used to create signature of a video. Here we will classify and compare existing video copy detection systems based on features they use. We mainly focus on visual features suitable for video copy detection, includes spatial features of keyframes, temporal features and motion features of video sequence. Spatial features of keyframes are categorized into global and local features.

#### A. Global Features

Global features provide invariant description of a video frames rather than using only selective local features. This approach works quite well for those video frames with unique and discriminating color values. Though merits are being easy to extract and require low computational cost but global features failed to differentiate between foreground and background. Global features are categorized as follows,

*1) Discrete Cosine Transform (DCT)*:

The essentiality of using image transformation is in removal of redundancy within neighboring pixels. Efficacy of a transformation scheme is laid in its ability to wrap up input data into as few transform coefficients as possible. This will allow quantizer to remove coefficients with small amplitudes without causing visual distortion in the reconstruction of an image. Due to DCT, most of the energy will be converged in lower level frequencies, so this will reduce the total amount of data that is required to describe an image or video frame. Yusuke et al. [10] perform feature extraction by applying 2D-DCT on each predefined block of keyframe to get AC coefficients, this DCT-sign based feature is used as signature of both reference and query video keyframes.

*2) Discrete Wavelet Transform*:

Gitto George Thampi et al. [11] use Daubechies wavelet transform to obtain feature descriptor from video frames. The wavelet coefficients of all frames of same segment are extracted and then mean and variance of the coefficients are computed to describe each segment of a video sequence.

*3) Ordinal Measure*:

Bhat et al. [12] used this feature for finding image correspondence to color degradation in original images, but it failed to remain robust against changes like rotation, flipping. This feature comprises ordered sequence of blocks of image based on their average intensity values. Xian-Sheng Hua et al. [13] use ordinal measure for generating signature of video segment, in which video frame is divided into number of blocks then for every block, average gray value is computed. Then these values are ranked in increasing order. The ranked sequence of average gray values gives ordinal measure; it incorporates rank–order of blocks of video frame according to their average gray values. It is highly invariant to color degradation but not to geometric transformations.

*4) GIST Feature*:

GIST feature represents abstract representation of a scene by extracting histogram of orientation gradients from fixed sized grids of video frame. GIST features have given good results in image classification, object recognition. Chenxia Wu et al. [14] use binarized form of GIST feature representation for each frame.

*5) Pyramid Histogram of Oriented Gradients (PHOG)*:

PHOG descriptor gives spatial pyramid representation of HOG descriptor. PHOG features are obtained by firstly extracting edge contours using canny edge detector for entire image. Then each image is divided into sub-regions at several pyramid level. PHOG descriptor represents each image sub-region with histogram of orientation gradients (HoG) at every resolution. Chenxia Wu et al. [14] use binarized form of PHOG feature representation for each frame by extracting binary PHOG feature for every frame with concatenation of all binarized HOG values from each sub-region of video frame. PHOG features have shown good results in object recognition.

*6) Color based Feature*:

Color based signature [15] has simple search routine but is sensitive to color shifts. Because color shifts is common attack for copying videos and color signatures will not work on black and white video content, most systems use the luminance component or grey-scale image in implementations.

TABLE I. CLASSIFICATION OF EXISTING VIDEO COPY DETECTION SYSTEMS BASED ON VISUAL FEATURES

| Feature Signature | Feature Type | Distance/Similarity Metrics and Search Mechanisms | Transformation Detection Invariance (Strengths) | Transformation Detection Variance (Weaknesses) | Improved Factors |
|---|---|---|---|---|---|
| 2D-DCT + BoVW [10] | Global | IDF weighting + Burstiness-aware scoring | T3-T6,T8,T10 | T1,T2,T7,T9 | Time Accuracy |
| Mean and variance of wavelet coefficients [11] | Global | Euclidean distance + Clustering based search | S3,S5,S6, S11,S13,S14 | S1,S9,S10,S15 | Accuracy |
| BPHOG + BGIST [14] | Global | Hamming distance + Copy confidence score | T1,T2,T4-T10 | T3 | Accuracy |
| MSF-color feature [15] | Semi-global | Edit distance based sequence matching | S1-S5,S9 S11-S15 | S6,S10 | Time |
| Spatial correlation descriptor [16] | Global | Chi-squared statistics + Edit distance | S1-S11, S14,S15 | S13 | Accuracy |
| BGH + IOM + SURF [17] | Global+Local | Hamming Embedding + Euclidean distance + Smith Waterman algorithm | T1, T3-T8 | T2,T9,T10 | Accuracy |
| SIFT [18] | Local | SVD + Graph based matching | T1-T10 | - | Accuracy |
| Hessian Laplace + CSLBP [19] | Local | Hamming Embedding + Hough Transform | T1-T10 | - | Accuracy |
| Hessian + CSLBP [20] | Local | K-nearest neighbor search + Hough Transform | T1-T4,T5-T7 | T8-T10 | Accuracy |
| MPEG-7 Motion Descriptor [24] | Motion | L1-norm Euclidean distance | T1,T2,T3, T6,T7 | T10 | Accuracy |
| Shot length sequence [25] | Temporal | Matching using suffix array structure | S1-S5,S9,S13-S15 | S6,S10,S11 | Time |
| SIFT + Ordinal measure [27] | Global+Local | Transformation adaptive matching | T1-T6,T8,T10 | T1,T7,T9 | Accuracy |

But luminance based methods perform poorly for transformations like cropping, zooming, text insertion, letter-box and pillar-box effects.

*7) Spatial Correlation Descriptor*:

Spatial correlation descriptor [16] uses inter-block relationship which encodes the inherent structure (pairwise correlation between blocks within video frame) forming unique descriptor for each video frame. The relationship between blocks of video frame is identified by content proximity. Original video and its transformed version will not be having similar visual features; however they preserve distinct inter-block relationship which remains invariant. This descriptor performs quite well for color changes and vertical deformations but failed to flip operation as this remodels graph structure of blocks in a video frame.

*8) Block-based Gradient Histogram(BGH)*:

The usage of global feature helps to enlist original video faster than local features, due to which retrieval speed gets improved significantly. Hui Zhang et al. [17] employs BGH which is to be extracted from set of keyframes. Firstly keyframes are divided into fixed number of blocks and for every block a multidimensional gradient histogram is generated. Set of these individual gradient histograms constitutes BGH feature for every keyframe. BGH is found to be robust against non-geometric transformations.

B. Local Features

Local feature based methods firstly identify points of interest from keyframes. These points of interest can be edges, corners or blobs. Once the interest point is chosen, then it is described by a local region surrounding it. A local feature represents abrupt changes in intensity values of pixel from their immediate neighborhood. It considers changes occurred in basic image properties like intensity, color values, texture. An interest point is described by obtaining values like gradient orientations from a region around that interest point. Local feature based CBCD methods [17,18,19,20] have better detection performance on various photometric and geometric transformations but only disadvantage is being high computational cost in matching.

*1) Scale Invariant Feature Transform(SIFT)*:

SIFT [21] employs Difference of Gaussian to detect local maxima values and these interest points are described by gradient histogram based on their orientations. Hong et al. [18] use SIFT descriptor due to its good stability and discriminating ability. SIFT feature performs well among local feature category and is robust to scale variation, rotation, noise, affine transformations.

*2) Speeded-Up Robust Features(SURF)*:

SURF [22] feature is based on Haar wavelet responses summed up around point of interest, which give maximum value for Hessian determinant. SURF is highly robust against geometric transformations like image scaling, translation, and rotation. Hui Zhang et al. [17] use SURF feature for representing points of interest having local maxima. SURF feature has better real time performance as compared to SIFT.

*3) Hessian-Laplace Feature*:

This feature is combination of Hessian affine detector and Laplacian of Gaussian. It employs Laplacian of Gaussian to locate scale invariant interest points on multiple scales. While at every scale, interest point attaining maximum value for both trace and determinant of Hessian matrix are selected to be affine invariant interest points. Hessian-Laplace is invariant to many transformations like scale changes, image rotation and due to detection is done at multiple scales so it is quite resilient to encoding, blurring, additive noise, camcording effects. Local feature based CBCD methods [19,20] employ Hessian-Laplace feature along with Center-Symmetric Local Binary Patterns (CSLBP) for feature description. CSLBP descriptor does not use color values so it is highly invariant to many photometric transformations.

*C. Motion Features*

Color based features have difficulty in detection of camera recorded copy as frame information gets significantly distressed. This problem can be efficiently resolved by employing motion features which use motion activity in a video sequence as it remains unchanged in severe deformations. Motion vectors have not been best choice for content based copy detection due to following reasons,

i) When motion activity is recorded at normal frame rate, it is almost zero, so it may not have any significant information.

ii) Motion vectors extracted at normal frame rate may appear to scatter in all directions due to inaccurate calculations as neighboring pixel values are close to each other in successive video frames.

iii) A static video content like news channel interview program does not have much motion to capture so motion vector value is near to small value or zero.

Tasdemir et al. [23] tried to solve above problems by lowering frame rate at which motion vectors are extracted. Due to this change, large sized vectors are obtained as motion activity between $1^{st}$ and 5th video frames is more than motion activity between consecutive frames. Tasdemir et al.[23] divide individual video frame into number of blocks and record motion activity between blocks of consecutive frames at reduced frame rate.

Roopalakshmi et al. [24] has implemented similar type of descriptor known as motion activity descriptor, for measuring activity of a video segment whether it is highly intense or not. This motion activity descriptor derives intensity of action, major direction of motion activity, distribution of motion activity along spatial and temporal domains.

*D. Temporal Features*

Temporal features represent variations in scene objects with respect to time domain rather than examining spatial aspect of each video frame. Shot length sequence [25] captures drastic change in consecutive frames of a video sequence. This sequence includes anchor frames which represent drastic change across consecutive frames. This sequence is computed by enlisting time length information among these anchor frames. Shot length sequence is distinctly robust feature as any separate video sequences will not be having set of successive anchor frames with similar time segment.

IV. PERFORMANCE EVALUATION

In performance evaluation mainly two measures have been employed, i) Normalized Detection Cost Rate (NDCR) combines cost of miss and cost of false alarm. Lesser NDCR value corresponds to better result. ii) F1 score considers both precision and recall to form harmonic mean, to assess copy localization accuracy of CBCD system. Higher F1 measure shows better performance. Table II shows performance of representative CBCD algorithms for different transformations of TRECVID (2008/2009/2011) dataset. Few observations can be made from this evaluation,

*1)* Local feature based CBCD algorithms [18,19,20] have shown better detection rate but extraction of local features along with their matching process have significant time requirements.

*2)* Video preprocessing done by CBCD systems [14,16,19,26] include removal of black border, picture-in-picture, camcording effects. Due to such preprocessing the global features can effectively deal with tough transformations.

*3)* Yusuke et al. [10] applied concept of bag-of-visual-words with DCT-sign based feature, which is usually used with local features. So applying concepts of local features with global ones can efficiently increase robustness of global features against various transformations.

*4)* As global features do not able to cope with geometric transformations, these global features can be efficiently combined with local features [17,27] to strengthen them against both photometric and geometric transformations.

TABLE II. PERFORMANCE OF REPRESENTATIVE VIDEO COPY DETECTION SYSTEMS FOR VARIOUS VIDEO TRANSFORMATIONS (T1-T10)

| Measures | NDCR (must be lower) | | | | | | F1 (must be higher) | | | | | |
|---|---|---|---|---|---|---|---|---|---|---|---|---|
| Datasets | TRECVID'08 | | | TRECVID'09 | | TRECVID'11 | TRECVID'08 | | | TRECVID'09 | | TRECVID'11 |
| Features | Local | Local | Local | Spatial | Global | Global | Local | Local | Spatial | Spatial | Global | Global |
| Methods | [18] | [19] | [20] | [27] | [10] | [14] | [18] | [19] | [17] | [27] | [10] | [14] |
| T1 | 0.12 | 0.079 | 0.224 | - | - | 0.881 | 0.94 | 0.948 | 0.68 | - | - | 0.958 |
| T2 | 0.13 | 0.015 | 0.321 | 0.58 | 1.0 | 0.687 | 0.94 | 0.952 | 0.4 | 0.72 | 0.0 | 0.943 |
| T3 | 0.14 | 0.015 | 0.079 | 0.23 | 0.007 | 0.470 | 0.90 | 0.950 | 0.8 | 0.94 | 0.977 | 0.958 |
| T4 | 0.15 | 0.023 | 0.064 | 0.41 | 0.000 | 0.448 | 0.93 | 0.946 | 0.82 | 0.84 | 0.967 | 0.958 |
| T5 | 0.07 | 0.000 | 0.023 | 0.32 | 0.000 | 0.284 | 0.95 | 0.949 | 0.84 | 0.88 | 0.961 | 0.949 |
| T6 | 0.11 | 0.038 | 0.064 | 0.24 | 0.000 | 0.425 | 0.94 | 0.950 | 0.8 | 0.85 | 0.976 | 0.952 |
| T7 | 0.12 | 0.065 | 0.140 | - | - | - | 0.92 | 0.941 | 0.72 | - | - | - |
| T8 | 0.11 | 0.045 | 0.437 | 0.44 | 0.843 | 0.590 | 0.94 | 0.950 | 0.7 | 0.92 | 0.883 | 0.949 |
| T9 | 0.17 | 0.038 | 0.693 | - | - | - | 0.93 | 0.951 | 0.64 | - | - | - |
| T10 | 0.23 | 0.201 | 0.537 | 0.52 | 0.821 | 0.575 | 0.95 | 0.946 | 0.68 | 0.82 | 0.847 | 0.950 |

*5)* Motion features [23,24] are used to distinguish videos but are not robust to text insertions or other occlusions which block the motion from being captured. Transformations involving rotation will change direction of motion vectors and give poor results.

*6)* In addition to robustness and discriminating abilities, the extracted feature vector should be compact enough to perform fast matching operation, as compact signature requires minimum storage space and performs similarity measurement in less computation time.

## V. CONCLUSION

We have presented an overview of recent advancements in content based video copy detection. The existing approaches have been illustrated with main focus on invariant features they employed for performing video copy detection. In order to deal with different photometric and geometric attacks, researchers have mainly focused on generating robust and unique video signatures. Features based on global, local, temporal, and motion aspects have been incorporated to tackle various types of attacks/deformations. It is inherently difficult to avert post production attacks, few algorithms have taken additional measures in form of preprocessing, combination of global and local features to deal with post production attacks. Although satisfactory efforts have been taken in designing robust video copy detection systems, market is still in need of more resilient and attack-invariant video copy detection system.